\documentclass{bmvc2k}

\usepackage{times}  
\usepackage{helvet} 
\usepackage{courier}  
\usepackage{graphicx} 
\usepackage{multirow}
\usepackage{booktabs}
\usepackage{xcolor}
\usepackage{amsmath}
\usepackage{amssymb}
\newcommand{\modelF}{Tensor Composition Net for Visual Relationship Prediction}
\newcommand{\modelS}{TCN-VRP}
\newcommand{\keypoint}[1]{\vspace{0.1cm}\noindent\textbf{#1}\quad}
\newcommand{\cut}[1]{}

\title{Tensor Composition Net for Visual Relationship Prediction}

\addauthor{Yuting Qiang}{qiangyuting.new@gmail.com}{1,2}
\addauthor{Yongxin Yang}{yongxin.yang@ed.ac.uk}{2}
\addauthor{Xueting Zhang}{Xueting.Zhang@ed.ac.uk}{2}
\addauthor{Yanwen Guo}{ywguo@nju.edu.cn}{1}
\addauthor{Timothy Hospedales }{t.hospedales@ed.ac.uk}{2,3}

\addinstitution{
Department of Computer Science and Technology \\ Nanjing University \\ Nanjing, China
}
\addinstitution{
School of Informatics \\ The University of Edinburgh \\ Edinburgh, UK
}
\addinstitution{
 Samsung AI Research Centre \\ Cambridge, UK
}

\runninghead{Qiang et.al }{Tensor Composition Net for Visual Relationship Prediction}

\def\eg{\emph{e.g}\bmvaOneDot}
\def\Eg{\emph{E.g}\bmvaOneDot}
\def\etal{\emph{et al}\bmvaOneDot}

\begin{document}

\maketitle

\begin{abstract}
We present a novel Tensor Composition Net (TCN) to predict visual relationships in images. Visual Relationship Prediction (VRP) provides a more challenging test of image understanding than conventional image tagging, and is difficult to learn due to a large label-space and incomplete annotation. The key idea of our TCN
is to exploit the low-rank property of the visual relationship tensor, so as to leverage correlations within and across objects and relations, and make a structured prediction of all visual relationships in an image. To show the effectiveness of our model, we first empirically compare our model with Multi-Label Image Classification (MLIC) methods, eXtreme Multi-label Classification (XMC) methods and VRD methods. We then show that, thanks to our tensor (de)composition layer, our model can predict visual relationships which have not been seen in training dataset. We finally show our TCN's image-level visual relationship prediction provides a simple and efficient mechanism for relation-based image-retrieval even compared with VRD methods.
\end{abstract}


\section{Introduction}
\label{Introduction}

Building on recent progress in image classification~\cite{kamavisdar2013survey} and object detection~\cite{ren2015faster}, understanding the high order semantics of complex images is becoming increasingly topical in machine perception. This requires not only recognizing individual objects, but also predicting visual relationships~\cite{Sadeghi2011RVP,lu2016visual}. Extracting visual relationships in images is critical to  comprehend the visual world beyond mere object classification.

\indent To this end, many recent works address \emph{Visual Relationship Detection (VRD)} ~\cite{lu2016visual,xu2017scenegraph,Zellers2018Motifs}. These works commonly formulate visual relationships as \emph{subject-predicate-object} triplets (e.g. \emph{person-ride-bike}, \emph{dog-next to-cat})~\cite{Zellers2018Motifs,xu2017scenegraph}. With the supervision of both the category and location (bounding box) of each object, they train or fine-tune an object detection network (e.g. Faster R-CNN) first, then a predicate prediction step is followed. In this paper, we promote the task of \emph{Visual Relationship Prediction (VRP)}, where visual relationships are predicted, but not registered to specific bounding boxes. Accordingly, our VRP problem only require image level visual relationship annotations without any localization.

Due to the lack of object level annotations (bounding boxe and object category), we could simply regard each unique visual relationship triple as a distinct category and formulate VRP as a multi-label image classification (MLIC) problem. However, one critical challenge is that the space of potential relationships is combinatorially large. There are $n\times n \times m$ potential relationships for a dataset with $n$ subjects, $n$ objects and $m$ predicates. A simple heuristic for managing this growth in label-space is to filter out visual relationships never seen during training. However the number of unique seen relationships is already large (over $20000$ for Visual Genome dataset), which is orders of magnitude larger than that of common MLIC datasets  ($20$ for PASCAL and $80$ for MSCOCO).

More fundamentally, this strategy prevents us from predicting unseen visual relationships at runtime. In many object classification problems the train and test dataset easily share exactly the same label space. But due to huge label space and long tail distribution of visual relationships, it is common to have testing relationships that are unseen during training. Therefore, an effective VRP model should be capable of predicting unseen/rarely seen relationships. Fortunately, different relationships are semantically related~\cite{lu2016visual}. For example, \emph{person-ride-horse} might be understandable without any training examples, given prior experience of ``person-ride-bike'' and ``horse'' object, if the model can re-combine the riding predicate and horse object.

\indent To enable such compositional reasoning, and to exploit correlations within- and across relationship triples, we model the space of relationships to predict as a three-way tensor instead of  a long vector. This provides two benefits: (i) It enables zero-shot prediction of novel relationships at runtime, (ii) it provides knowledge sharing between every occurrence of any given subject, object, or predicate among all unique relationships in which it occurs. \cut{Furthermore (iii) it can leverage correlations within and across relationships. E.g., detecting `person-drive-car` makes it more likely that `car-on-road`.} To achieve these benefits our model
  termed \modelF{} (\modelS) exploits the low rank property of the proposed tensor by incorporating a tensor composition layer into a neural network so that it can predict all potential relationships without inducing a combinatorial number of parameters. \cut{This enables \modelS{} to outperform other related MLIC and VRD methods on visual relationship prediction (especially for unseen/few-shot relationships) and image-query-by-relation.}


\indent To summarize our contributions: (i) We propose a simple and elegant neural tensor network to predict visual relationships in an image without filtering out unseen relationships. To the best of our knowledge, this is the first work to employ a neural tensor network to solve VRP based on image-level annotations; \cut{(ii) We empirically validate that, by employing GAP layer and fusing features of different scales, our \modelS{} can perform better on VRP tasks} (ii) Thorough experiments show that \modelS{} provides better visual relationship prediction compared to prior MLIC and XMC alternatives, especially for unseen or few-shot relationships. \cut{Thus it provides a more reliable approach to visual relationship retrieval even compared with VRD methods.} (iii) We show that TCN-based VRP provides a rich yet efficient query modality for image retrieval.

\section{Related Work}
\label{sec:Related_Work}
\keypoint{Visual Relationships}
\cut{As an intermediate level representation connecting images and language, understanding visual relationships has gained increasing attention in computer vision community over the past decade~\cite{Ramanathan2015semantic,Zellers2018Motifs,zhang2017vtranse}. \cut{A number of early works~\cite{Galleguillos2008co-occurrence,Gould2008,Kumar10efficientlyselecting,Russell2006MultiSeg} extract visual relationships using hand-crafted features to facilitate other tasks like image classification, retrieval or segmentation. Due to the lack of a generic visual relationship dataset, early works usually focused on a handful specific types of relationship like spatial relationships~\cite{Jia20133DReasoning,Silberman2012IndoorSeg,Zheng2015Scene} or human actions~\cite{Yao2010Grouplet,Ramanathan2015semantic}. In comparison, recognizing a wide variety of \emph{generic} visual relationships is a much more important yet challenging task.} With the release of visual relationship datasets~\cite{lu2016visual,VisualGenome} and the continued advance of object detection methods~\cite{girshickICCV15fastrcnn}, a growing body of work~\cite{zhang2017vtranse,xu2017scenegraph} has studied detecting \emph{generic} visual relationships. Due to the huge number of potential visual relationships, directly detecting each type of visual relationship in an image is infeasible. Instead, }
With the release of visual relationship datasets~\cite{lu2016visual,VisualGenome} and the continued advances of object detection methods~\cite{girshickICCV15fastrcnn}, a growing body of work~\cite{zhang2017vtranse,xu2017scenegraph,Zellers2018Motifs} has studied detecting \emph{generic} visual relationships. These methods usually decompose VRD task as object detection and predicate recognition so that the problem is tractable. Message passing technology (e.g. graph convolution, RNN) is used in these methods to boost both object detection and predicate recognition\cut{modules through structured prediction}. Although recently popular, these methods require RPNs, object detectors and  expensive message passing techniques for inference, which makes them complicated and computationally costly to apply in practice. Crucially, due to reliance on object detection, these models require high-quality annotations of object localization. \cut{which is less scalable than image-level annotation.}

In this paper, we propose to address the VRP rather than VRD task. Although VRP produces less detailed information (no bounding boxes), it only requires image-level annotation to train. Furthermore, one key motivating application of visual relationship detection is relationship-based image-retrieval. This particular capability does not depend on detection and can be driven by VRP, making it simultaneously simpler and cheaper to train.

\keypoint{Multi-Label Classification} Multi-Label Image Classification(MLIC) aims to annotate images with multiple labels. Early attempts~~\cite{boutell2004PR} tackle this problem by reformulating it as single-label classification.\cut{Several variants~\cite{raman2009ICML,cheng2010ICML} of this method have been proposed by considering inter-label dependency. However, for VRP, this group of methods is not applicable because it would require millions of binary classifiers.} Recent studies~\cite{li2016CVPR,wang2016cnnrnn,yang2016ECCV} pay more attention on label dependencies to facilitate multi-label prediction. Various strategies including exploiting graph structure \cite{li2016CVPR,chen2019CVPR}, matrix completion \cite{Cabral2015PAMI} and recurrent neural networks~\cite{wang2016cnnrnn} have been explored to model label correlations\cut{for MLIC}. However, many previous strategies can not be directly applied to VRP. For example, \cite{chen2019CVPR} use a predefined matrix to represent label co-occurrence, and information propagation is conducted by multiplying the feature map with this matrix. In our VRP context, the size of this matrix would be over $1e^{10}$, making it computationally intractable.

MLIC typically only considers dozens of object categories, another group of work focus on eXtreme Multi-label Classification (XMC), where the label space is extremely large (usually over 1 million).
These works usually employ label embeddings or tree to build correlation between different labels and make a structed prediction. For our VRP task, one crucial problem is to model the semantic correlation among different labels, which has not been considered by previous XMC methods.

\keypoint{Tensor Decomposition} Tensor decomposition methods are widely used in relationship induction for knowledge-bases \cite{ICML2011RESCAL,socher2013reasoning}\cut{, where the decomposition trains subject and object embeddings that combine with relationship matrices or core tensors to predict individual relationships}. Popular approaches include training a low-rank approximation of a label tensor to induce missing labels \cite{ICML2011RESCAL}, or training low-rank factors to estimate a label tensor~\cite{socher2013reasoning} using a neural network. These methods address completing the missing element in a single relationship tuple -- without any grounding to an image. In contrast, we use tensor composition to generate a unique image-specific label tensor that predicts all relations in an input image. Few studies have used tensor methods in the context of VRD. \cite{CVPR2018Tensorize} constructs a visual relationship tensor as prior to boost visual relationship detection~\cite{xu2017scenegraph}. It is completely different to ours in that we dynamically predict the tensor corresponding to each input image.
\cut{\cite{CVPR2018Tensorize} constructs a visual relationship tensor and combines it with the iterative inference method in \cite{xu2017scenegraph} to boost visual relationship detection. However this is completely different to ours in that it computes a single tensor as a prior, while we dynamically predict the tensor corresponding to each input image.}
\section{Method}
\label{Approach}
\keypoint{Problem Definition and Notation}
We assume a dataset of $N$ images, annotated with image-level visual relationships, based on $n$ object and $m$ predicate types. Each image is annotated with a number of relationship triplets, and a triplet is represented as
$\langle o_i, p_k, o_j \rangle$ to denote the fact that, in this image, the object $i \in \{1, 2, \ldots, n\}$ has the predicate $k \in \{1, 2, \ldots, m\}$ with the object $j \in \{1, 2, \ldots, n\}$. The set of relationship triplets can be naturally encoded as a 3-way relation tensor $\mathbf{T}\in \mathbb{R}^{n\times n \times m}$\footnote{Assuming that every object in an image can potentially take both subject and object role in a relation}. A tensor entry $\mathbf{T}_{ijk} = 1$ denotes the existence of a relationship triplet $\langle o_i, p_k, o_j \rangle$ in the image, otherwise $\mathbf{T}_{ijk}=0$. Our overall goal is to predict a relationship tensor that describes the full set of relationships that exists in a given image. That is, given a set of images and associated label tensors $(\mathbf{x}^{(i)},\mathbf{T}^{(i)})^N_{i=1}$, we want to learn a model $f(\cdot)$ that predicts the label tensor corresponding to an image $f(\mathbf{x}^{(i)})\rightarrow \hat{\mathbf{T}}^{(i)} \approx \mathbf{T}^{(i)}$.

\cut{
\begin{figure}[t]
\begin{center}
\includegraphics[width=0.98\linewidth]{figs/tensor_representation.pdf}
\end{center}
\caption{An image (top-left) is annotated with a set of visual relationship triplets (top-right), i.e. $\langle {\color{red}\text{subject}}, {\color{green}\text{predicate}}, {\color{blue}\text{object}} \rangle$. These triplets can be encoded as a relation tensor $\textbf{T}$ (bottom-left), and an entry $\textbf{T}_{i,j,k}$ being positive indicates the existence of a certain relationship triplet. The slice of ``ground'' (object) and the slice of ``on'' (predicate) are presented (bottom-right).} 
\label{fig:tensor_representation}
\end{figure}}

\keypoint{Computational Challenge}
A na\"ive predictor for $\mathbf{T}$ given $\mathbf{x}$ contains $d\times (n\times n \times m)$ parameters, where $d$ is the image feature dimension. In modern CNNs, $d$ is usually larger than $1000$, and for the commonly used benchmark dataset, $n$ or $m$ is at the scale of $100$. This means the number of parameters in predictor alone scales up to $1$ billion easily, which is $15$ times larger than the \emph{whole} ResNet-152. Clearly, the computational cost is prohibitive, and we need to find a lightweight proxy of $\mathbf{T}$ instead of predicting it directly.

\keypoint{Tensor Decomposition} Since $\mathbf{T}$ is inherently a tensor, a route to a more compact form is tensor decomposition. Tucker-decomosition \cite{kolda2009tensor} is a popular and safe choice for tensor decomposition. It decompose $\mathbf{T}$ as $\mathbf{T} = \mathbf{S} \times_1 A \times_2 B \times_3 C$. Here $\times_i$ denotes tensor-matrix dot product at the $i$th axis of tensor. $\mathbf{S}$ is the 3-way tensor sized $r_1\times r_2 \times r_3$ (called \emph{core} tensor), and $A$, $B$, $C$ are matrices sized $r_1\times n$, $r_2\times n$, $r_3\times m$, respectively (called factor matrices). $(r_1,r_2,r_3)$ is called Tucker rank of tensor, and they are hyper-parameters in our context.
Since we usually set $r_1\ll n$, $r_2 \ll n$, $r_3 \ll m$, the number of entities in $\{\mathbf{S}, A, B, C\}$ is much smaller than the $\mathbf{T}$, but we can reconstruct  $\mathbf{T}$ from $\{\mathbf{S}, A, B, C\}$ for the final prediction.
\cut{Unlike matrix decomposition $A=UV$, there are multiple ways to decompose a tensor. Popular choices are CP-~\cite{carroll1970analysis,harshman1970foundations} and Tucker-decomposition \cite{kolda2009tensor}, which generalizes CP for order-3 tensors.}
\cut{Since $\mathbf{T}$ is inherently a tensor, a route to a more compact form is tensor decomposition. Unlike matrix decomposition $A=UV$, there are multiple ways to decompose a tensor, and popular choices are CP-, Tucker- and TT-decomposition \cite{kolda2009tensor}. While the ideal decomposition is problem dependent, 3-way tensors have a nice property that CP decomposition can be seen as reduced forms of Tucker-decomposition (not necessarily true for order over 3). Thus we use Tucker decomposition as a safe choice.}

\keypoint{Knowledge Sharing} We can propose to predict $\{\mathbf{S}, A, B, $ $C\}$ then reconstruct $\mathbf{T}$ instead of predicting $\mathbf{T}$ directly. \cut{This has already reduced the number of parameters in predictor significantly.} A question that arises, is whether it is necessary to deduce all of $\{\mathbf{S}, A, B, C\}$ from the image feature?

\cut{To answer this question, we need to understand the meaning of each element in $\{\mathbf{S}, A, B, C\}$.} In Tucker-decomposition, $A$, $B$, or $C$ can be understood as word embedding matrices defining three \emph{distinct} latent spaces. The use of three distinct spaces is intuitive because: (i) The fact that A and B are not tied, i.e., $A\neq B$, reflects that words have different embeddings depending on whether they are subject or object \cite{tu2017context}. (ii) The fact that $r_1\neq r_2 \neq r_3$ reflects that we do not need the same capacity to encode the words from different vocabularies.

If we exhaustively compute the outer products of the columns of $A$, $B$, $C$, we get a stack of $r_1\times r_2 \times r_3$ tensors, each sized $n\times n \times m$, and the role of core tensor $\mathbf{S}$ is simply to rescale those $n\times n \times m$ tensors in order to reconstruct the original tensor. Based on this understanding, each column in the factor matrix encodes embedding of a specific word, and the core tensor contains the coefficients to recover the relationships of the given image with those basis.
\cut{each outer product corresponds to a certain \emph{graph basis},and the core tensor contains the coefficients to recover the scene graph~\cite{xu2017scenegraph} of the given image with those basis.}
Therefore, we predict the core tensor $\mathbf{S}$ from the given image only, and factor matrices $A, B, C$ are shared among all images, i.e., not conditioned on image features.

\subsection{Tensor Composition Network}
\begin{figure*}[!htbp]
\begin{center}
\includegraphics[width=0.98\linewidth]{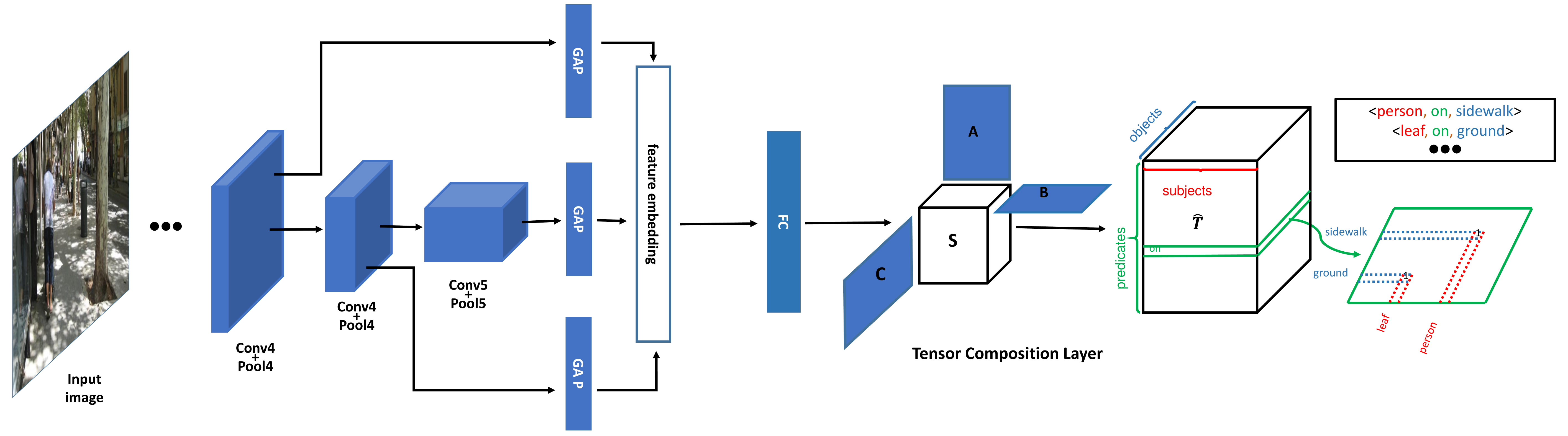}
\end{center}
\caption{Overview of our \modelF{} model.
An image (left) is to be annotated with relevant $\langle {\color{red}\text{subject}}, {\color{green}\text{predicate}}, {\color{blue}\text{object}} \rangle$ relationship triplets (right). These triplets can be encoded as a relation tensor $\hat{\textbf{T}}$, where a positive entry at $\hat{\textbf{T}}_{i,j,k}$ indicates the corresponding triplet exists. The tensor is generated via Tucker composition where the core tensor $\mathbf{S}$ is predicted from the image and contracted with three factor matricies $A,B,C$ that learn subject, predict and object embeddings. Blue blocks are learned parameters, and uncolored blocks (i.e., feature embedding, \textbf{S} and $\hat{\textbf{T}}$) are activations. 
}
\label{fig:overview}
\end{figure*}
\keypoint{Network Design}
Based on the reasoning above, we present our model -- Tensor Composition Network (TCN), illustrated in Fig.~\ref{fig:overview}. 
TCN consists of three components: (i) the image feature extractor $g_{\theta}(\mathbf{x})$ that produces a $d$-dimensional fused feature vector for a given image $\mathbf{x}$ (ii) the predictor that deduces the core tensor from image feature $\mathbf{S}=h_\phi(g_\theta(\mathbf{x}))$ (iii) the composer that generates the final prediction $t_{A,B,C}(S) = \mathbf{S} \times_1 A \times_2 B \times_3 C$. Here $h_\phi$ is a fully-connected layer that links every neuron produced by $g_\theta$ to every neuron in $\mathbf{S}$, which results in $d\times r_1 r_2 r_3$ parameters. The trainable parameters are $\{\theta, \phi, A, B, C\}$.

Similar to many related studies \cite{kamavisdar2013survey,ren2015faster}, we use a CNN to extract image features. To exploit features at different levels of abstraction that are helpful to predict relationships in scenes, we employ Global Average Pooling (GPA) to extract image features at multiple depths. The fused feature could enable the subsequent relationship prediction module to be aware of both local and global features.

With a fully connected layer, an intermedia vector is predicted from extracted image feature and it would be further folded as the core tensor (\textbf{S}). Finally, the visual relationship tensor is generated via Tucker composition where the core tensor $\mathbf{S}$ is contracted with three factor matricies $A,B,C$ that learn subject, predict and object embeddings.

\keypoint{Loss Function}
When we get the predicted tensor $\hat{\mathbf{T}}=t(h(g(\mathbf{x})))$,
\cut{we minimize its difference with the ground truth tensor $\mathbf{T}$. Mean Squared Error (MSE) is not the best choice for loss function because most of entities in $\mathbf{T}$ are zeros, the model tends to end up with a trivial solution -- it simply produces all-zero tensor for any input.

Instead of MSE, } we adopt to softmax~\cite{MICCV09TagProp} loss, which has been proven effective for multi-label prediction problem. Softmax loss is defined as,
\begin{equation}
\ell(\hat{\mathbf{T}}, \mathbf{T}) =  -\operatorname{sum}(\frac{\operatorname{flatten}(\mathbf{T})}{\operatorname{sum}(\mathbf{T})}\odot\log(\operatorname{softmax}(\operatorname{flatten}(\hat{\mathbf{T}}))))
\end{equation}

\noindent where $\odot$ denotes element-wise product. This is essentially a cross entropy loss with ground truth probability of  $\frac{\operatorname{flatten}(\mathbf{T})}{\operatorname{sum}(\mathbf{T})}$.

\cut{\keypoint{Initialization} To make our network easier and faster to train, we take advantage of the fact that we can estimate the factor matrices $A, B, C$ beforehand since they are set to be shared across instances. Specifically, we average all the $N$ relation tensors in the training set to get a global relation tensor $\bar{\mathbf{T}} = \frac{1}{N}\sum_{i=1}^{N}\mathbf{T}^{(i)}$. We then decompose this tensor using Higher-Order SVD (HOSVD)~\cite{DeSIAM00HOSVD} with a relative error threshold set for reconstruction, then factor matrices in decomposed $\bar{\mathbf{T}}$ can be used to initialize $\{A,B,C\}$. HOSVD also picks Tucker rank according to the relative error threshold, thus alleviates the need to set the hyper-parameters $r_1, r_2, r_3$ manually. }

\section{Experiments}
\label{Experiments}
As far as we know, this is the first work to address image level visual relationship prediction. To evaluate of our model, we  compare it with classic and state of the art MLIC and XMCmethods, with a particular focus on unseen and few-shot relationship prediction, and relation-based image retrieval (RBIR).

\keypoint{Datasets} 
VR~\cite{lu2016visual} and Visual Genome~\cite{VisualGenome} are commonly used benchmark datasets for scene understanding~\cite{xu2017scenegraph,zhang2017vtranse,Zellers2018Motifs}. As a small and early benchmark, VR only contains 5000 images with 37993 relationship instances in total. In contrast, Visual Genome contains more than 100k images but its raw annotations are very noisy, so data cleansing is needed before using it for visual relationship prediction. Among several cleaned version of this dataset~\cite{VisualGenome,xu2017scenegraph,zhang2017vtranse}, we use VG200~\cite{zhang2017vtranse} since it involves a relatively large number of images, object and predicate categories. We provide detailed statistics and low-rank analysis of the two datasets in supplemental material.


\keypoint{Implementation Details}
For fair comparison, we employ VGG-16~\cite{simonyan2014very} pretrained on ImageNet~\cite{ILSVRC15} as backbone network to extract features for all compared methods. We set relative error $\epsilon=0.02$ (VR) and $\epsilon=0.05$ (VG200) for HOSVD to initialize $A$, $B$, $C$.  Our model is trained for 40 epoches, using SGD with momentum, batchsize of 16 and an initial learning rate of $2\times 10^{-3}$, which is divided by 2 every 10 epoches. To address the problem of imbalanced distribution of visual relationship labels, we use weighted sampling during batch generation. Specifically, we first assign each existing visual relationship label $l_i$ ($i=1,2,\dots,L$) with a weight $w_{l_i}$, $w_{l_i} = \sum_{j=1}^{L}{t_{l_j}} / t_{l_i}$, $t_{l_i}$ denotes the number of times that the label $l_i$ occurs in the dataset. And the sampling weight for each image is $w_I =\sum_{l\in I}{w_l}$, so that images with rare relationship types or multiple relationships are prioritized during training.

\subsection{Visual Relationship Prediction}
\label{sec:VRP}
\keypoint{Setup} Given an image, we aim to predict the set of \emph{subject-predicate-object} triplets in it. Based on the predicted tensor $\hat{\mathbf{T}}$, we sort all potential triplets by their scores $\hat{\mathbf{T}}_{ijk}$ and select the top K triplets as our predicted relationships.

\keypoint{Metrics} As the annotated relationships in VR and Visual Genome are known to be incomplete~\cite{xu2017scenegraph,lu2016visual}, we follow previous works~\cite{xu2017scenegraph,Zellers2018Motifs,zhang2017vtranse} by using Top-K recall to measure performance. Specifically, we use $Recall@K$ ($K=20$, $50$ and $100$) and calculate the proportion of correctly predicted ground truth relationships in the top $K$ most confidently predicted relationships.

\keypoint{Competitors} We compare \modelS{} with both MLIC and XMC methods. Note that many state-of-the-art MLIC works employ a relational matrix to represent label dependencies which is not feasible for VRP due to the large label space, thus we compare: (1) WARP~\cite{Gong2013Deep}, which explores different loss functions for multi-label classification. We evaluate all three loss functions they consider (i.e. softmax~\cite{MICCV09TagProp}, pairwise ranking~\cite{pairwise2002Joachims} and weighted approximate ranking (WARP) losses). (2) Att-Consist~\cite{guo2019CVPR}, a two-branch network which introduces a new loss that measures the attention heatmap consistency between origin image and its transformed image to their network.\cut{We use their official code to perform visual relationship prediction task on benchmark datasets.} (3) CNN-RNN~\cite{wang2016cnnrnn} proposed a multi-label RNN model to sequentially predict the labels for each image, and the recurrent neurons in their model can capture high-order label co-occurrence dependency. For XMC competitors, we compare with 3 state-of-the-art methods : (4)Slice\cite{Slice}, a large-scale multi-label classification algorithm for  low-dimensional, dense, deep learning features. (5) ExMLDS~\cite{exmlds} leverage word embedding techniques for extreme multi-label classification. (6) Parabel~\cite{Parabel} promote XMC problem based on balanced label tree. For fair comparison, we employ the same feature extraction module as our \modelS{} for all XMC competitors.\cut{We implement CNN-RNN using the same architecture described in their paper, predicting one relation triple per iteration of the RNN.} (7) VTransE\cite{zhang2017vtranse} exploits embedding translation from the natural language processing to enable object-relation knowledge transfer and improve relationship detection. (8) VSPNet \cite{Zareian_2020_CVPR} proposed a bipartite message passing framework to detect visual relationships in an image. A graph-alignment algorithm is employed to enable the network to be trained without bounding-box annotation. Note that unlike the other competitors VTransE uses strong supervision (bounding boxes) during training, which provides an advantage.

\begin{table*}[t]
\begin{center}
\begin{tabular}{@{}rlcccccc@{}}
\toprule
\multirow{2}{*}{}& \multirow{2}{*}{Model} & \multicolumn{3}{c}{\textsc{VR}} & \multicolumn{3}{c}{\textsc{VG200}} \\
 \cmidrule(lr){3-5}  \cmidrule(lr){6-8}
 & & R@20& R@50 & R@100 & R@20& R@50 & R@100\\
\hline
\multirow{5}{*}{\rotatebox[origin=c]{90}{MLIC}}& SoftmaxLoss \cite{Gong2013Deep} & 23.73 & 36.26 & 46.31 & 30.32 & 43.32 & 50.85\\
& PairwiseLoss \cite{Gong2013Deep} & 24.94 & 37.56 & 46.91 & 30.72 & 45.05 & 56.41\\
& WARPLoss \cite{Gong2013Deep} & 23.70 & 34.18 & 41.93 & 30.28 & 43.23 & 52.94\\
& CNN-RNN \cite{wang2016cnnrnn}  & 23.80 & 35.90 & 44.94 & 26.13 & 39.67 & 50.74\\
& Att-Consist \cite{guo2019CVPR}  & 18.71 & 27.74 & 36.13 & \phantom{0}9.89 & 15.81 & 21.44\\
\midrule
\multirow{3}{*}{\rotatebox[origin=c]{90}{XMC}}&Slice\cite{Slice} & 23.29 & 32.26 & 39.69 & 29.43 & 40.12 & 44.37\\
& ExMLDS \cite{exmlds} & 16.61 & 23.16 & 27.08 & 22.45 & 32.21 & 38.46\\
& Parabel \cite{Parabel} & 24.31 & 29.48 & 34.64 & 28.77 & 39.43 & 43.35\\
\midrule
\multirow{2}{*}{\rotatebox[origin=c]{90}{VRD}} & VTransE$^{*}$ \cite{zhang2017vtranse} & 21.75 & 30.72 & 37.26 & 15.49 & 22.73 & 28.08 \\
& VSPNet \cite{Zareian_2020_CVPR} & - & - & - & 10.08 & 14.95 & 18.71 \\
\midrule
& Our \modelS{}\cut{-ws} & \textbf{29.42} & \textbf{43.54} & \textbf{53.46} & \textbf{32.97} & \textbf{47.82} & \textbf{59.04}\\
\bottomrule
\end{tabular}
\end{center}
\caption{Comparison of methods for relationship prediction. We re-evaluated Att-Consist~\cite{guo2019CVPR} using their official code but replacing the backbone with VGG16 newtork. Since CNN-RNN \cite{Gong2013Deep} and  \cite{wang2016cnnrnn} (WARP, Pairwise, Softmax) don't release their code, we implement their network according to their papers. $^*$ Strongly supervised.}
\label{Tab.Performance_on_PredCls_RelPred}
\end{table*}
\keypoint{Results}
Table~\ref{Tab.Performance_on_PredCls_RelPred} shows that our model is competitive with state-of-the-art alternatives. \cut{Crucially this is despite the fact that we make the prediction on the more challenging full set of possible relations rather than the filtered set of observed visual relations. E.g., \modelS{}'s chance value on VG200 is $1/4e^6$, while the MLIC methods' chance value is $1/20e^3$.}We attribute this success to our low rank Tensor composition layer, which compresses the label space and models label correlations with few learnable parameters. In comparison, CNN-RNN~\cite{wang2016cnnrnn} models
label correlation through sequential recurrence, outputting a label at each time-step given the previous steps' prediction. A wrong prediction at one step can affect all later predictions, a problem which is exacerbated in datasets such as VG200 with more relations per image. WARP~\cite{Gong2013Deep} introduces weighted approximate ranking loss to optimize top-k annotation accuracy, which means the advantage of WARP loss heavily rely on complete annotation for each image. However, acquiring a completely annotated VRP dataset is particularly expensive since there could be hundreds of relationships in each image. Att-Consist ~\cite{guo2019CVPR} addresses MLIC by penalizing the discrepancy between heatmaps of original image and transformed image in loss function. However, the increased complexity of optimization makes it easy to be trapped in a local minima.

VRD methods also provide potential competitors. In particular weakly-supervised VRD methods are fair competitors which use the same supervision (image-level triples) as the other competitors. Our results show that VRD methods do not necessarily perform well for our VRP problem. VSPNet \cite{Zareian_2020_CVPR} outperformed many alternatives on weakly supervised VRD. However, it  achieves comparatively low recall on our VRP problems. Our TCN can even outperforms the strongly supervised VRD method VTransE. This is because the VRD based methods heavily rely on the object detection module, If an object in the scene is missed by their detector (an event that is especially risky in the weak supervision case), VTransE and VSPNet can never recover. All relations involving that object are automatically missed.

\begin{figure}[t]
\begin{center}
\includegraphics[width=1.0\columnwidth ]{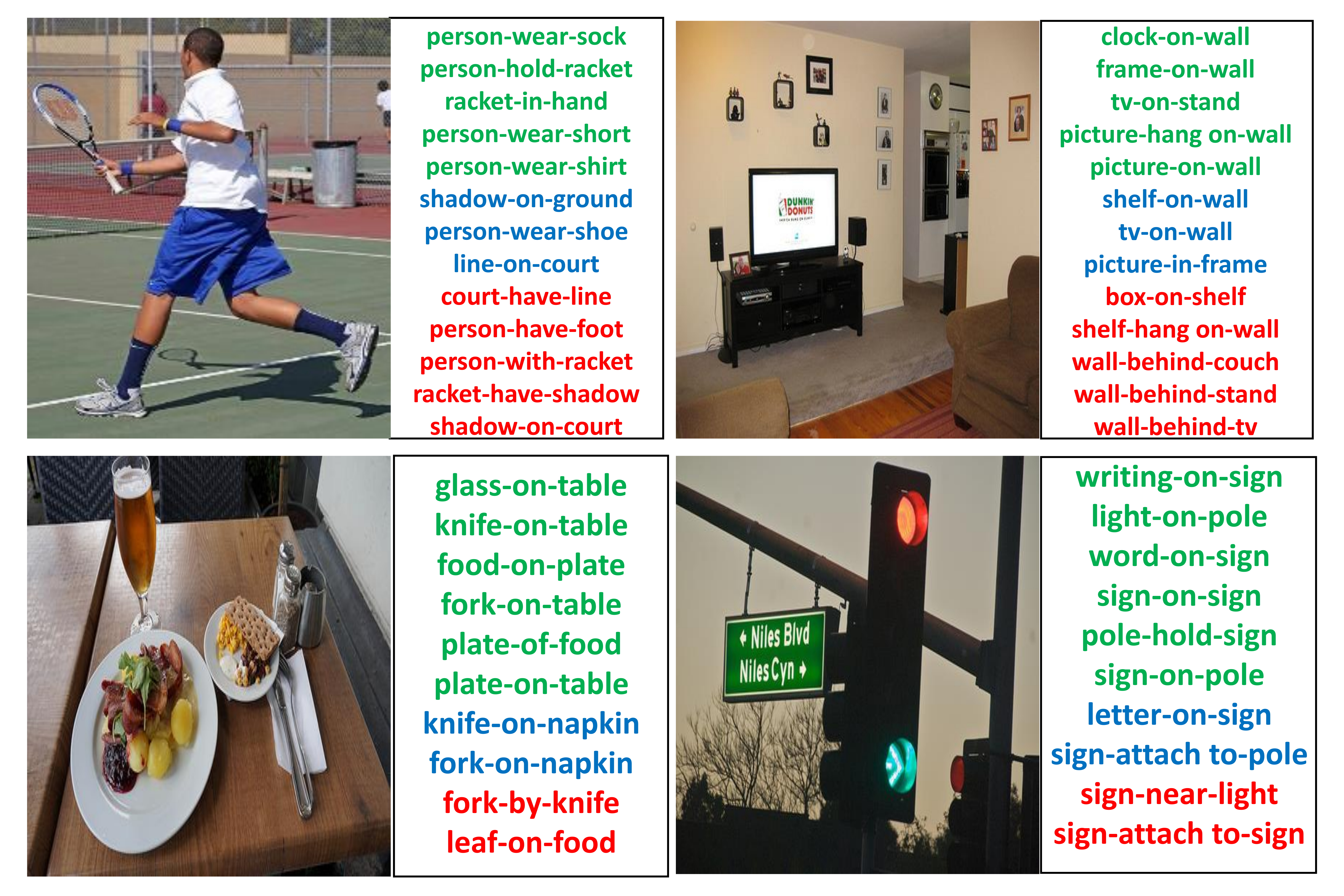}
\end{center}
\caption{Qualitative examples of \modelS{} relationship prediction using Recall@8. Sub-figure: Input image (left) and top 8 relationship predictions (right). Green: True positive predictions. Blue: False positive predictions. Red: False negative predictions.\cut{Despite mismatch to ground-truth, many of the false positive predictions are plausibly true.}}
\label{fig:relpred-qualitative}
\end{figure}

\keypoint{Qualitative Results}
Qualitative results of our model's relationship prediction  in Figure~\ref{fig:relpred-qualitative} suggest that our predictions are better than the quantitative results imply, since many predictions
that do not exist in the ground-truth (blue) are actually correct. On the other hand, small objects like ``leaf" in the bottom-left image are not recognized by our model.

\subsection{Few-shot and Zero-shot Relation Learning}
\label{sec:zero-shot learning}
Due to the long tail distribution and large label space, it is infeasible to build a VRP dataset with sufficient annotated instances for each relationship category. Some relationships may only occur few times or even never in the training dataset. Therefore, it is important for a VRP model to be able to predict visual relationships with few examples.

\keypoint{Setup, Metrics and Competitors} We compare the competing methods on $m$-shot ($m=0, 1, 5$) learning, by selecting those triplets which exist in the test dataset but occur $\leq m$ times in training. We use the sparser VR dataset for this experiment, and compare the same set of methods using $Recall@K$ evaluation as per general VRP. Existing VRD methods~\cite{liang2018visual, liao2019natural} tackling the zero-shot setting are trained under a strongly supervised condition and therefore are not directly comparable.



\keypoint{Results}
Table~\ref{Tab.few_shot_RelCls} presents results  on few-shot visual relationship prediction. XMC methods build correlations between different labels through label embeddings or a balanced tree, which helps them to perform better for few-shot relation prediction than MLIC methods. Our \modelS{} outperforms both MLIC and XMC methods by a large margin on few-shot learning, confirming that our tensor composition layer helps our model to learn from few examples. The top 100 recall of \modelS{} is at least 1.4 times better then the competitors. And the advantage is greater for 1 shot learning, while MLIC competitors achieve less than 1\% recall@100 for 1-shot relationship prediction, compared to our 6\% for this task.


\cut{CNN-RNN~\cite{wang2016cnnrnn} explores label dependencies to addreses multi-label classification and it achieves better performance on VG200 dataset compared to VR dataset. This is reasonable since images in VG200 dataset contains more relationships on average and label dependency can help predict few-shot relationships. However, our model can still outperform CNN-RNN on both dataset.}

\begin{table}[!htbp]
\begin{center}
\begin{tabular}{@{}rlccccccc@{}}
\toprule
\multirow{2}{*}{} & \multirow{2}{*}{Model} &  \multicolumn{2}{c}{0-shot} & \multicolumn{2}{c}{1-shot} & \multicolumn{2}{c}{5-shot}\\\cmidrule(lr){3-4}\cmidrule(lr){5-6}\cmidrule(lr){7-8}
& & R@50 & R@100 & R@50 & R@100 & R@50 & R@100\\
\hline
\multirow{5}{*}{\rotatebox[origin=c]{90}{MLIC}}& SoftmaxLoss\cite{Gong2013Deep}  & - & - & 0.37 & 0.81 & 4.35 & 7.96 \\
& PairwiseLoss\cite{Gong2013Deep} & - & - & 0.12 & 0.31 & 1.14 & 2.94 \\
& WARPLoss\cite{Gong2013Deep}  & - & - & 0.00 & 0.00 & 0.00 & 0.00 \\
& CNN-RNN~\cite{wang2016cnnrnn}  & - & - & 0.00 & 0.12 & 0.63 & 2.43 \\
& Att-Consist\cite{guo2019CVPR} & - & - & 0.00 & 0.00 & 0.00 & 0.00 \\
\midrule
\multirow{3}{*}{\rotatebox[origin=c]{90}{XMC}}& Slice\cite{Slice} & - & - & 2.04 & 3.84 & 6.90 & 12.11 \\
& ExMLDS\cite{exmlds} & - & - & 0.0 & 0.00 & 0.0 & 0.0 \\
& Parabel\cite{Parabel} & - & - & 2.11 & 4.28 & 4.63 & 8.43 \\
\midrule
& Our \modelS{} & \textbf{1.03} & \textbf{3.24} & \textbf{2.56}& \textbf{6.03} & \textbf{7.74} & \textbf{16.02}\\
\bottomrule
\end{tabular}
\end{center}
\caption{Few-shot visual relationship prediction results on VR. ``-'': Previous MLIC and XMC methods can't be applied to zero-shot visual relationship prediction. }
\label{Tab.few_shot_RelCls}
\end{table}
\subsection{Relation-Based Image Retrieval (RBIR)}
\label{sec:expr-triplet-retrieval}
A key benefit of visual relationship prediction is to enable image retrieval by more sophisticated relationship-triplet queries (e.g., \emph{person-push-bike}, \emph{dog-ride-surfboard})\cut{in contrast to the typical concept prediction-based queries which do not model visual relations between concepts \cite{wang2016cnnrnn}}. Recently, VRD-based methods have been applied to this task. We explore whether VRP is sufficient to perform RBIR well, given its appealing cheaper annotation requirements, and simpler inference method.

\keypoint{Setup and Metrics} Given a query triplet and a set of test images, we predict a relation score tensor for each image. We then sort all the images according to their scores for the query triplet. There could be multiple images containing the query triplet, so we regard an image as a correct hit if it has at least one query triplet in its annotation. We adopt the same setting as~\cite{zhang2017vtranse} which used Median rank (\textbf{Med r}) as metric. Median rank refers to the the median rank of the most confident correctly retrieved image. To show that our VRP model is sufficient for RBIR though it doesn’t localize relationships, we use the existing protocol of evaluating using the \emph{top 1000-frequent} triplets as queries~\cite{zhang2017vtranse}.\cut{We adopt the same setting as~\cite{zhang2017vtranse} which used Recall rate@5 and Median rank (\textbf{Med r}) as metric. Recall rate@k (\textbf{Recall@k}) computes the fraction of times a correct match was found among the top k retrievals and }
\cut{Note that in order to calculate the score for a specific triplet $\langle object-i, predicate-k, object-j \rangle$, we don't need to calculate the entire tensor. Instead we can calculate the triplet score efficiently as
\begin{equation}
\label{eq:triplet_score_computation}
\begin{split}
    T_{(i,j,k)} 
    & = S \times_1 A_{(i,:)}  \times_2 B_{(j,:)} \times_3 C_{(k,:)} \\
\end{split}
\end{equation}

\noindent where the rank reduction reduces the cost from $\mathcal{O}(rd^3)$ to $\mathcal{O}(r^3)$ with $
r\ll d$ .}

\keypoint{Competitors} To show the effectiveness of our \modelS{} for RBIR, we compare against VRD competitors. Specifically, we compare with VisualPhrase~\cite{Sadeghi2011RVP}, DenseCap~\cite{densecap}, LP~\cite{lu2016visual} and VTransE~\cite{zhang2017vtranse}. Note that these are all strongly supervised methods that benefit from box-level supervision during training that is not used by our TCN.
\cut{To simulate realistic circumstance, we first compare our model with previous state-of-the-art MLIC competitors~\cite{Gong2013Deep,guo2019CVPR} using \emph{random} selected 1000 triplets as query triplets (average training example per query 3 and 10 for VR and VG200). Additionally, }

\keypoint{Results} Tab.~\ref{Tab.single-triplet-retrieval} shows that our \modelS{} performs relation-based image retrieval comparably or better than state of the art VRD methods on both VR and VG200 benchmarks. One reason for this is the sparse annotation in the VR dataset, which makes it hard to finetune an accurate object detector on it, and that further affects the performance of VRD methods. In contrast, our \modelS{} is robust to sparse annotation and still achieves good performance on this dataset. Qualitative results of our model for RBIR are provided in supplemental material.

\begin{table}[!htbp]
\begin{center}
\begin{tabular}{@{}lcccccc@{}}
\hline
Dataset \textbackslash{} Method & VisualPhrase & DenseCap & LP & LP-VLK & VTransE & \modelS{}\\
\hline
VR \cite{lu2016visual} & 204  & 199  & 211 & 137 & \phantom{0}41& \phantom{0}\textbf{18}\\
VG200 \cite{zhang2017vtranse} & 18 & 13 & \phantom{0}- & \phantom{0}- &\phantom{0}7 & \phantom{0}\textbf{6}\\
\hline
\end{tabular}
\end{center}
\caption{RBIR (Relation-Based Image Retrieval) results versus VRD alternatives (Top-1000 frequent triplets). Metric is median rank ($\downarrow$) of the target image. \cut{: Recall@K ($\uparrow$) and Median Rank ($\downarrow$) metric.} Results for VisualPhrase \cite{Sadeghi2011RVP}, DenseCap \cite{densecap}, LP \cite{lu2016visual} and VTransE are reported in \cite{zhang2017vtranse}.\cut{ Methods are categorised by (D)etection or (P)rediction, which determines their annotation and runtime complexity. \modelS{} is very competitive, despite weaker supervision, and a simple detection-free inference procedure that is faster than VRD.}}\label{Tab.single-triplet-retrieval}
\end{table}
\subsection{Ablation Study}

We finally perform some further analysis to better understand our network. To anwser wheter Tucker factorization provides a good low-rank assumption to use as a layer in our model, we compare with a canonical polyadic (CP) alternative~\cite{kolda2009tensor} (\modelS{}-CP).  \cut{To answer this question, we compare with a canonical polyadic (CP) alternative \cite{kolda2009tensor}.} CP decomposes a three way tensor $\mathbf{T} \in \mathbb{R}^{d_1\times d_2 \times d_3}$ as a sum of R rank one tensors, i.e. $\mathbf{T} = \sum_{r=1}^{R} u_r^{(1)} \circ u_r^{(2)} \circ u_r^{(3)}$, where $\circ$ denotes outer product. We can predict all these $u_r^{(i)}$ (i=1,2,3) based on the extracted features and then perform CP tensor composition to construct the relation tensor $\mathbf{T}$. We then attempt to regard core tensor (i.e. $\mathbf{S}$) as parameter and predict factor matrics (i.e. $A, B, C$) for each image (\modelS{}-ABC). To evaluate the feature extraction strategies, \cut{We also ask how different feature extraction strategies affect the performance of our \modelS{}?}  We compare: (i) \modelS{}-FC. which takes the more common strategy \cut{e.g. Alexnet \cite{krizhevsky2014one} or VGG net~}\cite{simonyan2014very} of flattening the feature map of last max pooling layer.\cut{, and then using it to predict the core tensor with a fully connected layer.} (ii) \modelS{}-FL. Instead of just flattening the feature from last max-pooling layer, we use a GAP layer to embed the feature from last layer. \cut{This provides a simpler alternative to our multi-scale feature extraction strategy.}
\begin{table}[!htbp]
\begin{center}
\begin{tabular}{@{}lcccccc@{}}
\hline
\multirow{2}{*}{Method} & \multicolumn{3}{c}{VR} & \multicolumn{3}{c}{VG200}\\
 \cmidrule(lr){2-4}\cmidrule(lr){5-7}
 & R@20 & R@50 & R@100 & R@20 & R@50 & R@100 \\
\hline
TCN-VRP-CP  & 18.21 & 25.13 & 31.29 & 24.21 & 35.06 & 44.01\\
\modelS{}-ABC & 19.78 & 26.93 & 32.64 & 27.67 & 39.59 & 48.72 \\
\modelS{}-FC & 26.70 & 39.00 & 48.03 & 30.25 & 45.32 & 55.76\\
\modelS{}-FL & 27.86 & 41.01 & 50.87 & 30.05 & 43.80 & 54.57\\
\modelS{} & 29.43 & 43.54 & 53.46 & 32.97 & 47.82 & 59.04\\
\hline
\end{tabular}
\end{center}
\caption{Ablation study of our \modelS{} model}
\label{Tab:Ablation study}
\end{table}

Results in Table~\ref{Tab:Ablation study} show that: (i)  Tucker is much better than CP-composition for relationship prediction. Our Tucker decomposition model explicitly separates common knowledge about object/predicate from specific information about an image, which makes it  more reliable and stable to train. (ii) Core tensor ($\textbf{S}$), rather than factor matrics ($A, B, C$) should be regarded as image-specific. This corroborates our earlier analysis about Tucker-decomposition for visual relationship tensor. \cut{that \modelS{} works by locating each image ($\textbf{S}$) within a graph basis ($A, B, C$).}(iii) The GAP feature extraction helps our model resist over-fitting by extracting a lower dimensional representation. (iv) Our \modelS{} can be further improved by fusing features from different pooling layers to combine global and local information.

\section{Conclusion}
We proposed a neural tensor network to predict visual relationships in images. By introducing a
Tensor composition layer, we compressed the relation tensor label-space and leverage semantic correlations between relationships. We achieve better visual relationship prediction compared
to related MLIC methods. We also achieve competitive relation-based image-retrieval performance with strong supervised VRD works. In future we will explore scaling annotation by using existing caption data together with text parsing to automatically acquire weak relation annotation at large scale.

\bibliography{references}   
\clearpage
\begin{appendix}
\maketitle
\section{supplemental material}
\subsection{Datasets and Low Rank Assumption}
Detailed statistics of the VRD and VG datasets are reported in Table \ref{Tab.datastat}. Our TCN-VRP relies on the assumption that relation tensors can be approximated by a set of low-rank factors. To validate this assumption, we employ HOSVD to decompose the average global tensor $\bar{\mathbf{T}}$ with different reconstruction errors. The results in Table~\ref{Tab.error_vs_ranks} show that 2\% reconstruction error can be achieved using less than half the original input dimensions, which corresponds to compressing the tensor to less than 7\% of original size.
\begin{table}[h]
\begin{center}
\resizebox{1.0\columnwidth}{!}{
\begin{tabular}{@{}rllccc@{}}
\hline
Dataset & \#Train img. & \#Test img. & \#Avg Rels &  \#Objs & \#Preds \\
\hline
VRD & \phantom{0}4,000 & \phantom{0}1,000 &  \phantom{0}7.6 & 100 & \phantom{0}70 \\
VG200 & 73,801 & 25,857 & 11.8 & 200 & 100 \\
\hline
\end{tabular}
}
\end{center}
\caption{Statistics of different datasets. The number of train images, test images, relationships per image (on average), object categories and predicate categories are shown. }
\label{Tab.datastat}
\end{table}

\begin{table}[!htbp]
\begin{center}
\resizebox{1.0\columnwidth}{!}{
\begin{tabular}{@{}lcccc@{}}
\hline
Dataset & & \multirow{2}{*}{$\epsilon=0.02$} & \multirow{2}{*}{$\epsilon=0.05$} &  \multirow{2}{*}{$\epsilon=0.10$} \\
\multicolumn{2}{l}{\hspace{-0.25cm}(Sub,Obj,Pred) Dimensions} \\
\hline
VRD & Tucker Rank: & 27, 26, 17 & 12, 10, 10 & 6, 4, 6\\
(100, 100, 70) & Compression: & 0.026 & 0.006 & 0.002\\
VG200 & Tucker Rank: & 94, 73, 33 & 55, 36, 14 & 23, 16, 5\\
(200, 200, 100) &  Compression:& 0.066 & 0.012 & 0.003\\
\hline
\end{tabular}}
\end{center}
\caption{Representing relation tensor with Tucker composition. Input tensor dimension (Subj,Obj,Pred), Tucker rank and compression ratio, at various reconstruction error thresholds.}
\label{Tab.error_vs_ranks}
\end{table}

\subsection{Qualitative results of Relation-Based Image-retrieval(RBIR)}
\begin{figure}[!hbpt]
\begin{center}
\includegraphics[width=0.99\linewidth]{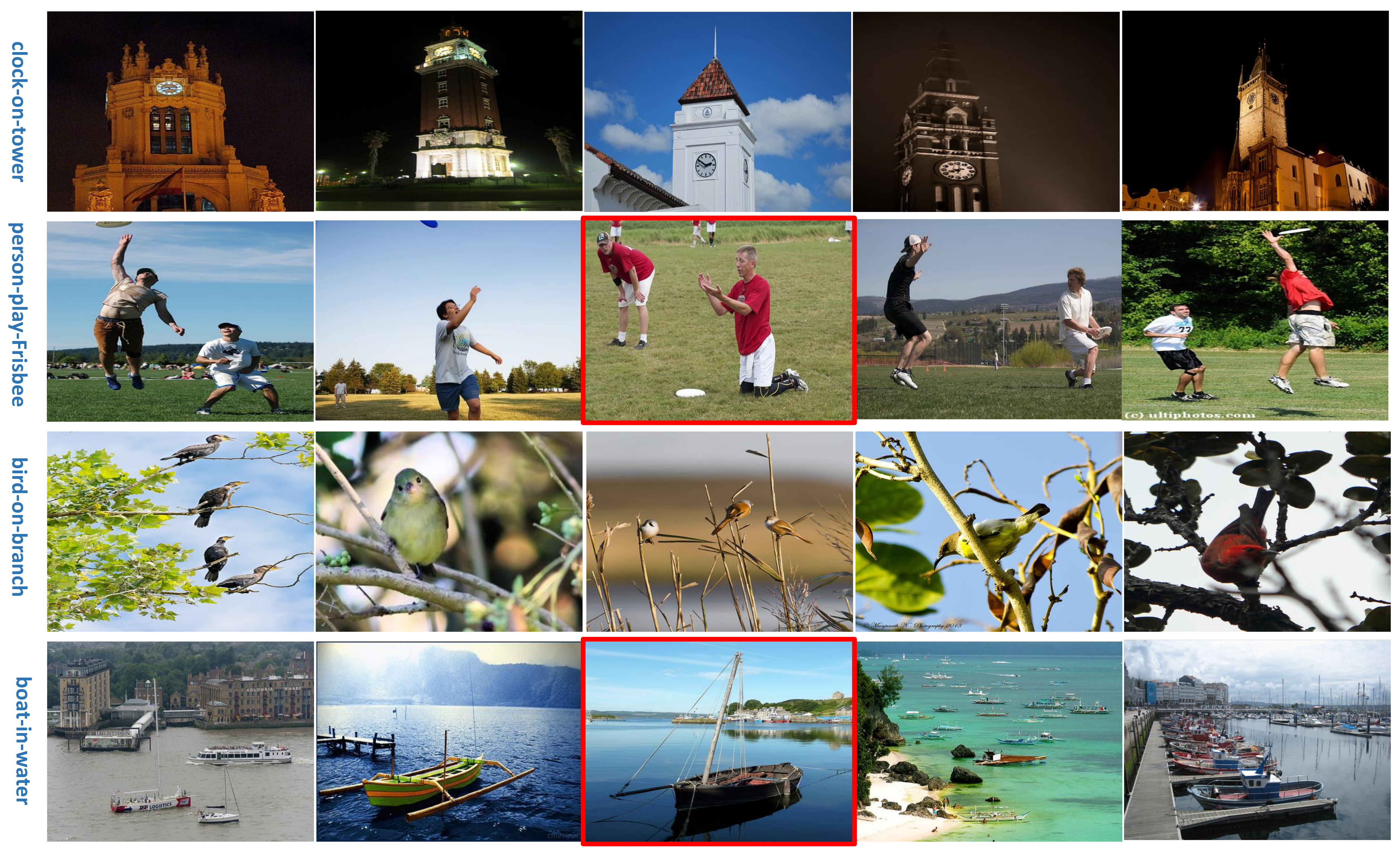}
\end{center}
\caption{Qualitative examples of relation-based image-retrieval.  The four rows (from top to bottom) show Top 5 results for: \emph{clock-on-tower}, \emph{person-play-frisbee}, \emph{bird-on-branch} and \emph{boat-in-water}, respectively. Red frames are false positives. The image in last row is tagged with \emph{boat-on-water} rather than \emph{boat-in-water}, but it should be regard as correct.}
\label{fig:single-triplet-retrieval}
\end{figure}

To qualitatively evaluate our model for relation-based image-retrieval, we use four different triplets (i.e. ``clock-on-tower", ``person-play-Frisbee", ``bird-on-branch" and ``boat-in-water" ) as image retrieval queries, as shown in Figure \ref{fig:single-triplet-retrieval}. For ``clock-on-tower" and ``bird-on-branch", our  top five returned images match the query exactly. As for `` person-play-Frisbee", our model retrieved a wrong image (the third  of second row) since the person is not ``playing" Frisbee although  ``person" and ``Frisbee" objects exist. Another wrong retrieval result is the third ranked result for ``boat-in-water". The image is tagged instead with ``boat-on-water", but this should also be regarded as a correct retrieval given ``boat-in-water''.
\end{appendix}
\end{document}